\documentclass[letterpaper, 10 pt, conference]{ieeeconf}  

\IEEEoverridecommandlockouts                              
\usepackage[utf8]{inputenc}
\usepackage[T1]{fontenc}

\usepackage{cite}
\usepackage[pdftex]{graphicx}
\usepackage{amsmath}
\usepackage{amssymb}
\usepackage{array}
\usepackage{wasysym}
\usepackage{bm}
\usepackage{siunitx}
\usepackage{url}
\usepackage{color}
\usepackage{epstopdf}
\usepackage[ruled]{algorithm2e} 
\usepackage{arydshln}
\usepackage{mathtools}

\usepackage{stfloats}
\usepackage{relsize}
\usepackage{cancel}
\usepackage{epstopdf}
\usepackage{environ}
\usepackage{amsthm}

\NewEnviron{ruledtable}{%
\par\addvspace{2mm}\hrule height 0.03cm 
\begin{table}[!h]\BODY\end{table}
\hrule height 0.03cm \addvspace{2mm}
}

\DeclareMathOperator*{\argmin}{arg\,min}

\theoremstyle{plain}

\newtheorem{definition}{Definition}

\usepackage{CJKutf8}

\renewcommand{\baselinestretch}{0.98}

\begin{document}
\title{\LARGE \bf{Safety-critical Control of Quadrupedal Robots with Rolling Arms \\ for Autonomous Inspection of Complex Environments}}
\author{Jaemin Lee, Jeeseop Kim, Wyatt Ubellacker, Tamas G. Molnar, and Aaron D. Ames
\thanks{This work is supported by Dow under the project  \#227027AT.}
\thanks{$^{1}$ The authors are with the Department of Mechanical and Civil Engineering, California Institute of Technology (Caltech), Pasadena, CA, USA,
{\tt\small \{jaemin87, jeeseop, wubellac, tmolnar, ames\}@caltech.edu}}%
}

\maketitle

\begin{abstract}
This paper presents a safety-critical control framework tailored for quadruped robots equipped with a roller arm, particularly when performing locomotive tasks such as autonomous robotic inspection in complex, multi-tiered environments.  In this study, we consider the problem of operating a quadrupedal robot in distillation columns, locomoting on column trays and transitioning between these trays with a roller arm. 
To address this problem, our framework encompasses the following key elements: 1) Trajectory generation for seamless transitions between columns, 2) Foothold re-planning in regions deemed unsafe, 3) Safety-critical control incorporating control barrier functions, 4) Gait transitions based on safety levels, and 5) A low-level controller. Our comprehensive framework, comprising these components, enables autonomous and safe locomotion across multiple layers. We incorporate reduced-order and full-body models to ensure safety, integrating safety-critical control and footstep re-planning approaches. We validate the effectiveness of our proposed framework through practical experiments involving a quadruped robot equipped with a roller arm, successfully navigating and transitioning between different levels within the column tray structure.
\end{abstract}

\section{Introduction}
\label{section1}
Legged robots have emerged as invaluable assets in unstructured and hazardous environments, capitalizing on their exceptional agility. Within the industrial sector, legged robots have assumed tasks previously relegated to human labor, such as autonomous facility inspections and maintenance tasks \cite{gehring2021anymal, halder2023construction}. Industrial facilities, often characterized by multiple tiers or levels, such as column trays \cite{moran2017distillation}, introduce a unique navigational challenge, necessitating seamless and safe transitions between these levels. Our previous research developed a robotic system equipped with a roller arm and rigorously evaluated its performance through high-fidelity simulations \cite{molnar2023mechanical}. Building upon this work, this paper presents a safety-critical control framework tailored to our robotic system, primarily focusing on ensuring safety within the column tray. This comprehensive framework spans every facet of the robotic control pipeline, encompassing trajectory generation, safety-critical control, footstep re-planning, gait transition, and low-level feedback control to ensure safety while the robot is operated in the confined space.

\subsection{Related Work}
Quadruped robots have harnessed agile locomotion capabilities, enabling them to excel in traversing rough terrains and uneven surfaces \cite{raibert2008bigdog,fankhauser2018robust,kim2020vision, miki2022learning, ubellacker2023robust}. The stability and robustness inherent in quadrupedal locomotion make it suitable for tasks such as carrying payloads in unstructured environments \cite{tournois2017online, dadiotis2022trajectory} and negotiating obstacles while executing predefined missions \cite{gilroy2021autonomous, gaertner2021collision,lee2023hierarchical}. The integration of perception and vision-based algorithms has further augmented the locomotion capabilities of quadruped robots \cite{miki2022learning,kim2020vision,jenelten2020perceptive,grandia2023perceptive}. While these studies focus on enhancing the productive and efficient locomotion of quadruped robots on a single level, the challenge of completing missions in multi-layer columns, requiring transitions between levels, remains unaddressed.

\begin{figure}[t] 
\centering
\includegraphics[width=0.9\linewidth]{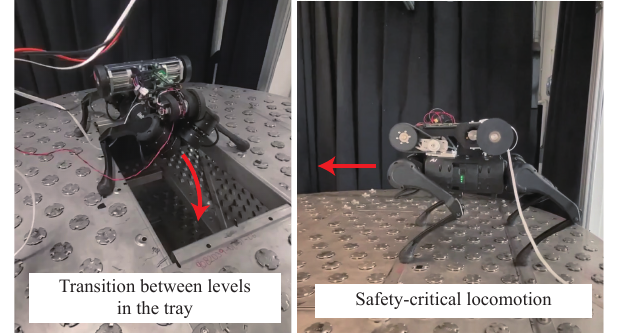}
\vspace{-2mm}
\caption{\textbf{Key functionalities of robots in a column tray}: It is imperative to incorporate essential functionalities, including the implementation of seamless transitions between levels and the assurance of safety-critical locomotion within the column.}
\label{Fig1}
\vspace{-0.8cm}
\end{figure}

Efforts to enhance the functionality of quadruped robots have seen the introduction of robotic arms \cite{ferrolho2022roloma,sleiman2023versatile} or upper-body structures \cite{kashiri2019centauro} mounted atop the robot's base to perform locomanipulation tasks. However, these robotic systems have not been specifically engineered to operate within multi-layered environments, necessitating essential functionalities, as illustrated in Fig. \ref{Fig1}. In our prior work, we meticulously designed the mechanical structure of the robot to enable its operation within column layers, substantiating its performance through high-fidelity simulations \cite{molnar2023mechanical}. Conventional control frameworks for legged robots typically consist of multiple layers, encompassing walking pattern generation utilizing reduced-order models and full-body feedback control to embody the walking pattern \cite{lee2023hierarchical,gehring2013control, kim2019highly, bellicoso2018dynamic} and whole-body control schemes empower robots to dynamically perform a hierarchical tasks \cite{lee2012intermediate, lee2020mpc}, including agile locomotion and maintaining robust balance through solid contacts \cite{kim2020dynamic,lee2022online}. However, these research efforts do not provide unified frameworks tailored for safe operation in multi-layered environments, as we propose in this paper.

Control Barrier Functions (CBFs) have emerged as a fundamental tool for ensuring the safety of robot operations, with a growing body of research demonstrating their effectiveness \cite{agrawal2017discrete,khazoom2022humanoid,lee2023hierarchical,kim2023safety}. CBF-based control methods have found wide-ranging applications, such as guiding bipedal robots in determining secure footstep placements \cite{agrawal2017discrete} and averting self-collisions in humanoid robots \cite{khazoom2022humanoid}. Additionally, CBFs have been instrumental in resolving conflicting safety constraints within quadruped robots \cite{lee2023hierarchical} and enforcing holonomic constraints among various robot components \cite{kim2023safety}.

Despite the practicality and intuitiveness of CBF-based approaches employing reduced-order models, it is well-recognized that simplified model-based methods may not fully capture the feasibility of full-order (full-body) systems \cite{lee2020reachability}. Consequently, there arises a need to incorporate additional re-planning mechanisms to ensure that the outcomes of CBF-based methods remain feasible and reliable for real-world robotic applications. Furthermore, to the best of our knowledge, the application of CBFs to dynamically adjust walking gaits while simultaneously ensuring the robot's safety has not been extensively explored. 

\subsection{Contributions}
\begin{figure}[t] 
\centering
\includegraphics[width=\linewidth]{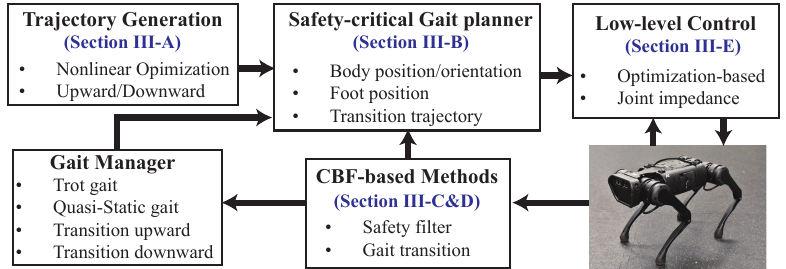}
\vspace{-5mm}
\caption{\textbf{Overview of the proposed layered architecture}: The proposed framework comprises several distinct layers: trajectory generation, a safety-critical gait planner, methods based on CBFs, and a low-level controller.}
\label{Fig2}
\vspace{-0.7cm}
\end{figure}

This paper introduces a pioneering multi-layered architectural framework designed for the operation of quadruped robots within column layers. Our proposed architecture harnesses various key components, including optimization-based trajectory generation, safety-critical control, foothold re-planning strategies, and low-level full-body control, as illustrated in Fig. \ref{Fig2}. We employ CBFs configured in an ellipsoidal form to safely control the robot's base and facilitate gait transitions, such as transitioning between trot and quasi-static gaits. Our hypothesis advocates for gait transitions as the robot approaches unsafe regions, enhancing navigation safety. Furthermore, we introduce a foothold re-planning strategy aimed at relocating unsafe footsteps into the safe region by determining the position that minimizes the distance from the originally planned one.

The primary contributions of this paper can be summarized as follows: First, we improve the trajectory generation method to ensure safe transition behaviors between levels. Second, we guarantee safety and avoid unsafe areas by simultaneously leveraging the CBF-based safety filter using a reduced-order model and the foothold re-planning with a full-body model in the multi-layer architecture. Finally, we utilize CBFs to make the gait transition while the robot safely does locomotion in the column. All components of our approach are validated through rigorous hardware experiments.

The structure of our paper is as follows. In Section \ref{section2}, we elucidate the models utilized in this study and provide essential background information on safety-critical control with reduced-order models. Section \ref{section3} delineates the comprehensive layered architectural framework, encompassing trajectory generation for transitioning between levels, foothold re-planning, CBF-based safety-critical control, gait transition strategies, and low-level controllers. Finally, Section \ref{section4} presents the experimental results that demonstrate the practical application of our approach on a quadrupedal robot platform operating within a layered environment.

\section{Preliminaries}
\label{section2}
In this section, we briefly overview the models utilized in our framework, including both full-order and reduced-order models. In addition, we introduce the fundamental concept of CBFs that underpins our approach.

\subsection{Full-order and reduced-order models}
For legged robots, the rigid-body dynamic equation is described by considering a configuration space $\mathcal{Q}\subset \mathbb{R}^{n}$ and an input space $\mathcal{U} \subset \mathbb{R}^{n-6}$: 
\begin{equation} \label{rigid}
    \mathbf{D}(\bm{q}) \ddot{\bm{q}} + \mathbf{H}(\bm{q}, \dot{\bm{q}}) = \mathbf{S}^{\top} \bm{\tau} + \mathbf{J}_c(\bm{q})^{\top} \bm{F}_{c}
\end{equation}
where $\bm{q} \in \mathcal{Q}$ and $\bm{\tau} \in \mathcal{U}$ represent the joint variable and the control torque input, respectively. $\mathbf{D}: \mathcal{Q} \to \mathbb{S}_{++}^{n}$, and $\mathbf{H}:\mathcal{Q}\times \mathbb{R}^{n} \to \mathbb{R}^{n}$ represent the functional mappings for the mass/inertia matrix and the sum of Coriolis/centrifugal and gravitational forces, respectively. $\mathbf{S}\in \mathbb{R}^{(n-6)\times n}$ and $\bm{F}_c \in \mathbb{R}^{n_c}$ are the selection matrix for the actuation joints and the contact forces, where $n_c$ denotes the dimension of the contact forces under a certain supporting phase. $\mathbf{J}_{c}(\bm{q}) \in \mathbb{R}^{n_c \times n}$ is the contact Jacobian associated with the contact force. The following explicit equality constraints are enforced to maintain solid contact between the feet and the ground: 
\begin{equation} \label{holonomic}
\begin{split}
    \mathbf{J}_{c}(\bm{q}) \dot{\bm{q}} &= \bm{0},\quad \mathbf{J}_{c}(\bm{q}) \ddot{\bm{q}} + \dot{\mathbf{J}}_{c} (\bm{q}, \dot{\bm{q}})\dot{\bm{q}} = \bm{0}
\end{split}
\end{equation}
where $\mathbf{J}_c(\bm{q})$ is full-row rank. Based on the rigid-body dynamics of the robots \eqref{rigid} and multiple inequality and equality constraints, including the above holonomic constraints \eqref{holonomic}, the low-level feedback controllers can be designed as whole-body controllers or output feedback controllers. In this paper, we describe the full-body model and holonomic constraints in state-space form:
\begin{equation}
    \dot{\bm{x}}_{q} = f_{q}(\bm{x}_q) + g_{q}(\bm{x}_q) \bm{u} , \quad \psi_c(\bm{x}_q, \dot{\bm{x}}_q) = 0
\end{equation}
where $\bm{x}_q = [\bm{q}^{\top}, \dot{\bm{q}}^{\top}]^{\top}$ and $\bm{u} = [\bm{\tau}^{\top}, \bm{F}_c^{\top}]^{\top}$, respectively.

In the locomotion control and safety-critical re-planning steps, many reduced-order models have been frequently utilized, such as the inverted pendulum model, the lumped mass model, the double-integrator model, etc. In this paper, we select a double-integrator system to capture the kinematic behavior of the robots with a state $\bm{\varphi} = [\varphi_x, \varphi_y] \in \mathbb{R}^{2}$ in $2$-dimensional operational space, and then the equation of motion can be written as:
\begin{equation}
    \dot{\bm{x}}_{\varphi} \coloneqq \left[\begin{array}{c}  \dot{\bm{\varphi}} \\ \ddot{\bm{\varphi}} \end{array}\right] = \underbrace{\left[\begin{array}{cc} \bm{0} & \mathbf{I}  \\ \bm{0} & \bm{0}\end{array} \right]\left[\begin{array}{c}  \bm{\varphi} \\ \dot{\bm{\varphi}} \end{array}\right]}_{f_{\varphi}(\bm{x}_{\varphi})}  + \underbrace{\left[ \begin{array}{c} \bm{0} \\ \mathbf{I} \end{array} \right]}_{g_{\varphi}(\bm{x}_{\varphi})} \bm{\nu}
\end{equation}
where $\bm{\nu} \in \mathbb{R}^{2}$ denotes the input of double-integrator system. This reduced-order model is leveraged to implement a locomotion control process to generate control velocity commands and footholds associated with the commanded velocity. In addition, we use the double-integrator system to avoid unsafe areas by referring to our safety-critical control. 

\subsection{Control Barrier Function}
To utilize Control Barrier Functions (CBFs) for the reduced-order system defined in the previous section, we introduce a safe set $\mathcal{C} \subseteq \mathbb{R}^{4}$, which is the $0$-superlevel set of a continuously differentiable function $h:\mathbb{R}^{4} \to \mathbb{R}$, where $\mathcal{C} \coloneqq \{\bm{x}_{\varphi} \in \mathbb{R}^{4}: h(\bm{x}_{\varphi}) \geq 0 \}$. Given a state space $\mathcal{X}_{\varphi}\subseteq \mathbb{R}^{4}$ and an input space $\mathcal{U}_{\varphi} \subseteq \mathbb{R}^{2}$, we define a control barrier function according to \cite{ames2016control}.
\begin{definition}
    Given a safe set $\mathcal{C}$, $h$ is a control barrier function (CBF) if there exists a function $\alpha \in \mathcal{K}_{\infty}^{e}$ such that for all $\bm{x}_{\varphi} \in \mathcal{C}$: 
    \begin{equation*}
        \sup_{\bm{\nu} \in \mathcal{U}_{\varphi}} \dot{h}(\bm{x}_{\varphi}, \bm{\nu}) = \sup_{\bm{\nu} \in \mathcal{U}_{\varphi}} [\mathcal{L}_{f} h(\bm{x}_{\varphi}) +\mathcal{L}_{g} h(\bm{x}_{\varphi})\bm{\nu}] \geq - \alpha(h(\bm{x}_{\varphi}))
    \end{equation*}
where $\mathcal{L}_{f}h(\bm{x}_{\varphi}) = \frac{\partial h}{\partial \bm{x}_{\varphi}}(\bm{x}_{\varphi}) f(\bm{x}_{\varphi})$ and $\mathcal{L}_{g}h(\bm{x}_{\varphi}) = \frac{\partial h}{\partial \bm{x}_{\varphi}}(\bm{x}_{\varphi}) g(\bm{x}_{\varphi})$ are Lie derivatives. If all states in $\mathcal{C}$ satisfied the above inequality for inputs in $\mathcal{U}_{\varphi}$, $\mathcal{C}$ is  forward invariant. 
\end{definition}
\noindent
The above-defined CBF can be leveraged to obtain a safe control input by solving a simple quadratic programming (QP) problem. Given a designed control input $\bm{\nu}^{d}$, we formulate a least-square error minimization problem to compute the safe control input:
\begin{equation} \label{CBF_QP}
    \argmin_{\bm{\nu}\in \mathcal{U}_{\varphi}} \| \bm{\nu}^{d} - \bm{\nu} \|^{2} \quad \textrm{s.t.} \quad \dot{h}(\bm{x}_{\varphi}, \bm{\nu}) \geq - \alpha(h(\bm{x}_{\varphi})).
\end{equation}
In this paper, we employ this CBF-based safety filter to adjust the control input to ensure the safety of the robot in the locomotion process. 

\begin{figure}[t] 
\centering
\includegraphics[width=\linewidth]{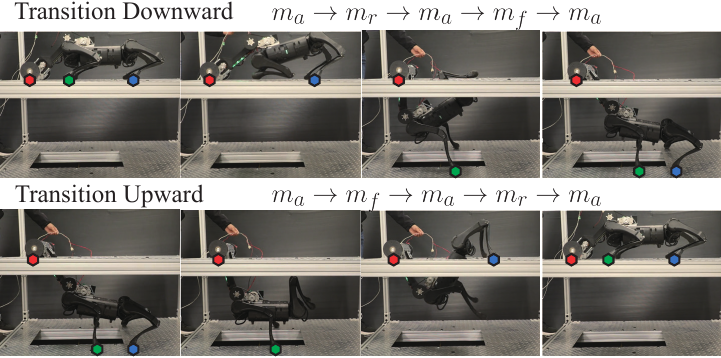}
\vspace{-5mm}
\caption{\textbf{Phases for transition behaviors}: In contact sequences, contacts with the roller arm are indicated in red, the front legs in green, and the rear legs in blue through corresponding markers.}
\label{Fig3}
\vspace{-0.7cm}
\end{figure}

\section{Multi-layered Architecture}
\label{section3}
This section introduces our proposed framework designed to enable the essential functionalities of the robot within the column tray, as depicted in Fig. \ref{Fig1}. We outline our methods, ranging from trajectory generation to the low-level controller described in Fig. \ref{Fig2}.

\subsection{Trajectory generation for transition behaviors}
The transition behaviors involve moving both downward from the upper level to the lower one and upward in the reverse direction. These transition motions consist of three contact phases: 1) all feet in contact ($m_{a}$), 2) rear feet in contact ($m_r$), and 3) front feet in contact ($m_f$). In our previous work \cite{molnar2023mechanical}, we proposed one candidate optimization problem to generate a joint trajectory. Based on the previous formulation, we have reformulated the trajectory generation problem and obtained a solution to the problem using FROST \cite{hereid2017frost}. In this context, we define a state $\bm{x}(t)\coloneqq [\bm{q}(t)^{\top}, \dot{\bm{q}}(t)^{\top}, \ddot{\bm{q}}(t)^{\top}, \bm{u}(t)^{\top} ]^{\top}$ given a finite time horizon $t\in [0,T]$ with discretization time interval $\Delta t$ ($t_i = i\Delta t$ where $T = N\Delta t$). Then, the problem is described as follows: 
\begin{equation}
    \begin{split}
        \min_{\bm{x}(0), \cdots, \bm{x}(T)} &\quad \sum_{i=0}^{N} \mathcal{J}(\bm{q}(t_i), \dot{\bm{q}}(t_i), \bm{u}(t_i)) \\
        \textrm{s.t.} & \quad \dot{\bm{x}}_q(t_i) = f_{q}(\bm{x}_q(t_i)) + g_q(\bm{x}_{q}(t_i)) \bm{u}(t_i),\\
        &\quad \psi_{c}(\bm{x}_q(t_i), \dot{\bm{x}}_q(t_i)) =0, \quad \bm{F}_c(t_i) \in \mathcal{FC},\\
        &\quad \bm{\varphi}_{\textrm{com}}(t_i) \in \mathcal{S}_{c}, \quad \psi_{\textrm{inq}} (\bm{x}_q(t_i)) \geq 0,\\
        &\quad \bm{q}(t_i) \in \mathcal{Q}, \quad \dot{\bm{q}}(t_i) \in \mathcal{V}, \quad \bm{\tau}(t_i) \in \mathcal{U},\\
        &\quad \bm{q}(0) \in \mathcal{Q}_0, \quad \bm{q}(T) \in \mathcal{Q}_{T}, \\
        &\quad \dot{\bm{q}}(0) = \dot{\bm{q}}(T) = 0
    \end{split}
\end{equation}
where $\mathcal{FC}$ and $\mathcal{S}_c$ denote the contact wrench cone and the supporting polygon in terms of the contact phases ($m_a$, $m_r$, $m_f$). For the implementation, we slightly modify the supporting polygon to prevent a large deviation of the center of mass (COM) from the initial position. In addition, $\bm{\varphi}_{\textrm{com}} \in \mathbb{R}^{2}$ represents the position of the COM on the horizontal plane. The state inequality function $\psi_{\textrm{inq}}$ is introduced to avoid collisions between the robot's legs, arm, and the surrounding environment. We further define $\mathcal{Q}_0\subset\mathcal{Q}$, $\mathcal{Q}_{T}\subset\mathcal{Q}$, and $\mathcal{V}\subset\mathbb{R}^{n}$ as the sets representing the initial and final robot configurations, as well as the velocity space, respectively. To be more specific, the sets $\mathcal{Q}_0$ and $\mathcal{Q}_{T}$ are determined based on geometric parameters of the manway, including its width $d_w$, the height $d_h$, and the spacing between the levels $d_{L}$.

Since the transition behaviors are composed of three different phases ($m_a$, $m_r$, $m_f$), it is essential to model the robotic system as a hybrid dynamical system with identity maps connecting the phases. For the implementation, we assign time duration for each phase: $t_a$, $t_r$, and $t_f$, such that their sum, denoted as $T$, depends on the transition type (downward/upward). As shown by Fig.~\ref{Fig3}, for downward transitions, the phase sequence unfolds as follows: ($m_{a} \to m_{r} \to m_a \to m_{f} \to m_{a}$). Conversely, for upward transitions, it follows the reverse sequence: ($m_a \to m_f \to m_a \to m_r \to m_a$). To optimize computational efficiency, we solve the formulated optimization problem using a relatively large time interval ($\Delta t = 0.1$ seconds). Subsequently, we generate trajectories by interpolating the computed waypoints using cubic splines ($\Delta t = 0.001$ seconds). 

\subsection{Footstep re-planning near unsafe region}
We employ a rectangular representation to delineate an unsafe region, referred to as the "manway", which the robot must avoid during locomotion. To determine whether an expected foothold $\bm{x}_f \in \mathbb{R}^{2}$ falls within or outside this unsafe region, we rely on the coordinates of its four vertices denoted as $\mathbb{V} = \{\bm{v}_1, \; \bm{v}_2, \; \bm{v}_3, \; \bm{v}_4 \}$. Each vertex $\bm{v} = [ v^{x},\; v^{y}]$ is characterized by its $x$ and $y$ positions, denoted by $v^{x}$ and $v^{y}$, respectively. The convex hull, derived from these vertices, is defined as follows:
\begin{equation}
\begin{split}
    \textrm{hull}(\mathbb{V}) \coloneqq \{ \bm{x}\in \mathbb{R}^{2}:& 0 \leq \vec{\bm{v}}_{12}\cdot (\bm{x} - \bm{v}_{1}) \leq \vec{\bm{v}}_{12} \cdot \vec{\bm{v}}_{12}, \\
    & 0 \leq \vec{\bm{v}}_{23} \cdot (\bm{x} - \bm{v}_{2}) \leq \vec{\bm{v}}_{23} \cdot \vec{\bm{v}}_{23} \} 
\end{split}
\end{equation}
where $\vec{\bm{v}}_{ij} = \bm{v}_j - \bm{v}_i$ for $i,\; j \in \{1,\; 2,\; 3, \;4\}$. If the planned foot position falls within $\textrm{hull}(\mathbb{V})$, it indicates that the planned foothold either resides within the unsafe region or exactly on its boundary lines. In such instance, it is imperative to replace the initially planned foothold with a safe and feasible alternative. 

\begin{figure}[t] 
\centering
\includegraphics[width=\linewidth]{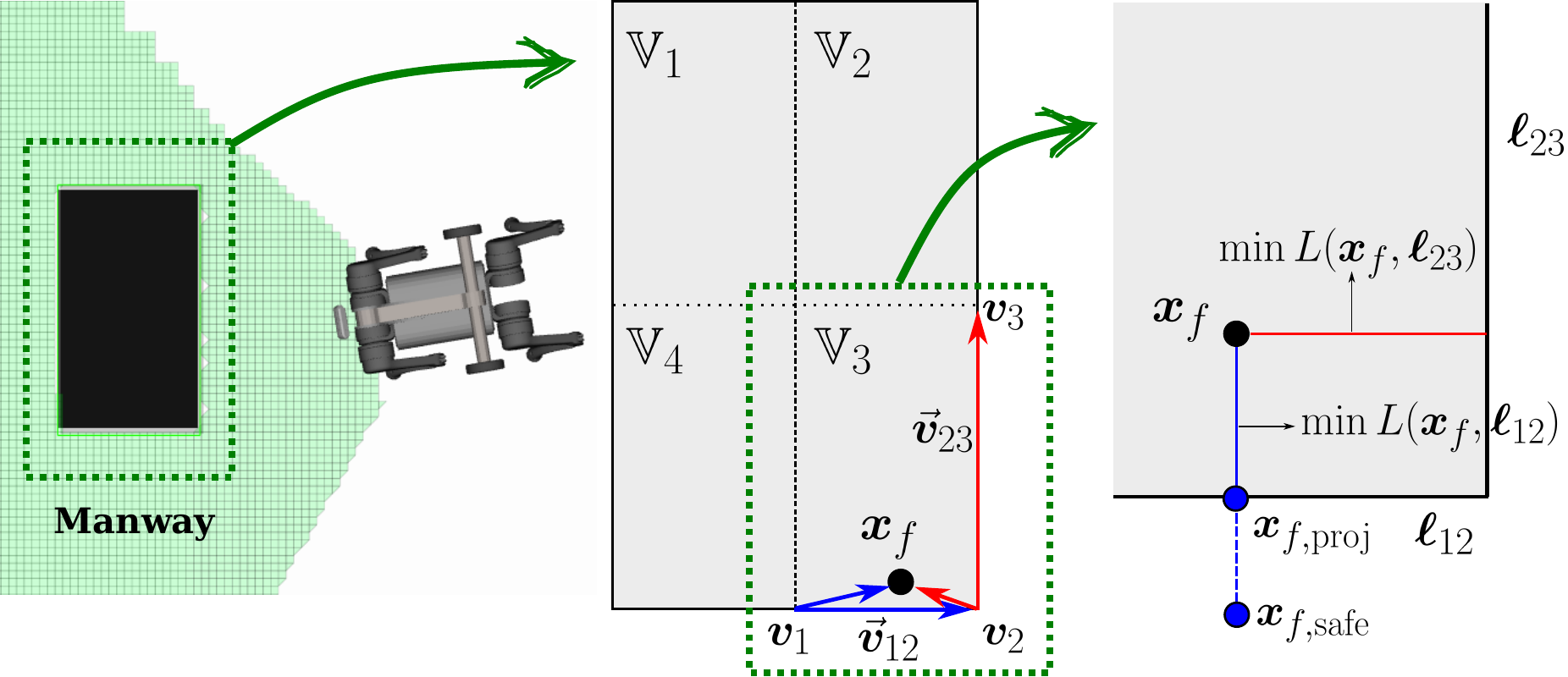}
\vspace{-5mm}
\caption{\textbf{Foothold re-planning in unsafe region}: We utilize geometric information to identify the nearest safe point near the planned foothold.}
\label{Fig4}
\vspace{-0.7cm}
\end{figure}

In our initial step, we partition the rectangular region into sub-regions, each sharing a common point within the rectangle. Specifically, in this paper, we use four sub-regions denoted as $\textrm{hull}(\mathbb{V}_1)$, $\textrm{hull}(\mathbb{V}_2)$, $\textrm{hull}(\mathbb{V}_3)$, and $\textrm{hull}(\mathbb{V}_4)$ where $\textrm{hull}(\mathbb{V}) = \cup_{i\in\{1,2,3,4\}} \textrm{hull}(\mathbb{V}_{i})$ (see Fig. \ref{Fig4}). Through iterative checks, we ascertain whether the planned foothold is situated within a specific hull $\bm{x}_{f} \in \textrm{hull}(\mathbb{V}_{s})$. Subsequently, we can determine the nearest safe location to the initially planned foothold using the following approach:
\begin{equation} \label{distance_min}
    \begin{split}
        \bm{x}_{f,\textrm{proj}}=\argmin_{\bm{x}} &\: ( \min_{\bm{x}}  \: L(\bm{x}_f, \bm{\ell}_{ij} ), \: \min_{\bm{x}} \: L(\bm{x}_f, \bm{\ell}_{jk}) )
    \end{split}
\end{equation}
where $L(\bm{x}_{f}, \bm{\ell}_{ij})$ represents the distance between the planned foothold $\bm{x}_{f}$ and a point $\bm{x}$ on a line $\bm{\ell}_{ij}$ where $\bm{v}_i$ and $\bm{v}_{j}$ both lie on the line $\bm{\ell}_{ij}$. In addition, $\bm{v}_{j}$ is the vertex that is distinct from other hulls, $\bm{v}_{j} \notin \mathbb{V}_{\{1,2,3,4\} \backslash s}$, and $\bm{v}_i$ and $\bm{v}_{k}$ are adjacent vertices to $\bm{v}_{j}$. 

Once the solution to the problem \eqref{distance_min} is obtained, we make a slight adjustment to position this point slightly further inside the designated safe region, enhancing the safety of the foothold.
\begin{equation}
    \bm{x}_{f, \textrm{safe}} = \bm{x}_{f} + (1 + \epsilon) ( \bm{x}_{f,\textrm{proj}} - \bm{x}_{f}) 
\end{equation}
where $\epsilon>0$ represents the weighting factor used to adjust the new foothold position further inside the designated safe region. If the swing foot has the capability to reach the re-planned foothold, we compute the control commands based on $\bm{x}_{f,\textrm{safe}}$. However, if the swing foot cannot attain this point, we consider an alternative point projected onto the other line to assess its kinematic feasibility for the swing foot.

\subsection{Safety-critical control using the reduced-order model}
\begin{figure}[t] 
\centering
\includegraphics[width=0.9\linewidth]{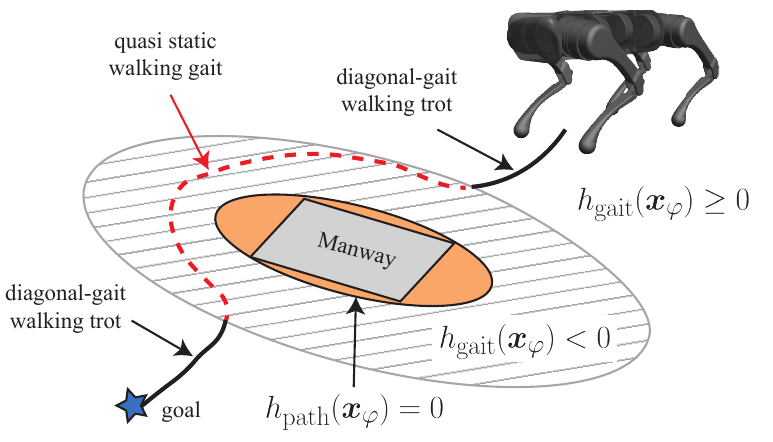}
\vspace{-4mm}
\caption{\textbf{Two CBFs for safe path and gait transition}: We have devised two Control Barrier Functions (CBFs) for distinct purposes. The first CBF is engineered to guide the robot within the safe region, while the second CBF serves as a trigger mechanism for gait type transitions.}
\label{Fig5}
\vspace{-0.7cm}
\end{figure}

In the context of safety-critical control, the detailed formulation of CBF is crucial, taking into account the geometry of obstacles or, in this case, the manway. Typically, CBFs are formulated in circular or linear forms to prevent collisions with obstacles or ensure safe behaviors. However, in scenarios where the unsafe region has a rectangular shape, such as the manway within the column layer in Fig.~\ref{Fig5}, we design the CBF using an ellipse equation, as outlined below:
\begin{equation}
\begin{split}
      h(\bm{x}_{\varphi}) =& (\bm{x}_{\varphi} - \overline{\bm{x}}_{\varphi})^{\top}\mathbf{A}(\bm{x}_{\varphi} - \overline{\bm{x}}_{\varphi}) - 1\\
      =& \bm{x}_{\varphi}^{\top} \mathbf{A} \bm{x}_{\varphi} + \mathbf{B}\bm{x}_{\varphi} + c\\
\end{split}
\end{equation}
where $\overline{\bm{x}}_{\varphi}$ denote a center of the ellipse. In detail, 
\begin{equation*}
    \begin{split}
      \mathbf{A} =& \left[\begin{array}{cc}\frac{\cos^2\theta }{a^2} + \frac{\sin^2\theta}{b^2} & \left(\frac{1}{a^2} - \frac{1}{b^2}\right)\cos \theta \sin \theta \\ \left(\frac{1}{a^2} - \frac{1}{b^2}\right)\cos \theta \sin \theta & \frac{\sin^2\theta }{a^2} + \frac{\cos^2\theta}{b^2} \end{array} \right] \\
      \mathbf{B} =& - 2 \overline{\bm{x}}_{\varphi}^{\top} \mathbf{A}, \quad  c = \overline{\bm{x}}_{\varphi}^{\top} \mathbf{A} \overline{\bm{x}}_{\varphi} -1
    \end{split}
\end{equation*}
where $a$, $b$, and $\theta$ are the lengths of the major and minor radii and the orientation of the ellipse. The lower bound of CBF is expressed explicitly: 
\begin{equation}
\begin{split}
    \dot{h}(\bm{x}_{\varphi}, \bm{\nu}) =& (2\mathbf{A} \bm{x}_{\varphi} + \mathbf{B}^{\top})^{\top}(f_{\varphi}(\bm{x}_{\varphi}) + g_{\varphi}(\bm{x}_{\varphi}) \bm{\nu}) \\
    \geq&  -\alpha(\bm{x}_{\varphi}^{\top} \mathbf{A} \bm{x}_{\varphi} + \mathbf{B}\bm{x}_{\varphi} + c).
\end{split}
\end{equation}
Considering the manway dimensions, which are the width $d_{w}$ and the height $d_{h}$, $a$ and $b$ can be determined as $a = \beta \frac{d_w}{2}$ and $b = \beta\frac{d_h}{2}$ where $\beta$ is a scaling factor aimed at making the ellipse more conservative. 

In this work, we employ CBFs to accomplish two primary objectives: 1) modify the robot's path around the manway ($h_{\textrm{path}}$), 2) change the robot's walking gaits ($h_{\textrm{gait}}$) in terms of the safety levels by selecting different scaling factors $\beta_{\textrm{path}}$ and $\beta_{\textrm{gait}}$. Let suppose that $h_{\textrm{path}}$ is designed and the desired path is given $\bm{x}_{\varphi}^{d}(t)$ for all $t \in [0,T]$, we can obtain the safe control input $\bm{\nu}_{\textrm{safe}}$ by solving the optimization problem \eqref{CBF_QP} with the desired control input $\bm{\nu}^{d} = \mathbf{k}(\bm{x}_{\varphi}, \bm{x}_{\varphi}^{d})$ where $\mathbf{k}: \mathcal{X}_{\varphi}\times \mathbb{R}^{4} \to \mathcal{U}_{\varphi}$ is the desired controller, which is locally Lipschitz continuous. The obtained control input $\bm{\nu}_{\textrm{safe}}$ is then applied by the locomotion controller with a specific gait.

\subsection{Locomotion control and gait transition}
In this section, we introduce two locomotion gaits, which are a diagonal-gait walking trot $\Xi_{\textrm{trot}}(H, \dot{\varphi}_{x}, \dot{\varphi}_{y}, \dot{\vartheta}_{z}) $ and quasi-static walking gait $\Xi_{\textrm{static}}(H, \dot{\varphi}_{x}, \dot{\varphi}_{y}, \dot{\vartheta}_{z})$ where $H$, $\dot{\varphi}_{x}$, $\dot{\varphi}_{y}$, and $\dot{\vartheta}_{z}$ are body height, linear velocity in $x$ and $y$, and angular velocity about the $z$ axis. The diagonal-gait walking trot is implemented with the Raibert heuristics, such as what we previously implemented in \cite{ubellacker2023robust}. For the quasi-static walking gait, only one leg moves during the swing phase and is supported by the remaining three legs. The walking gait is composed of four phases, $w_{\textrm{FL}}$, $w_{\textrm{BR}}$, $w_{\textrm{FR}}$, and $w_{\textrm{BL}}$ whose the swing foot is front left, back right, front right, and back left. We repeat iterating the phases in a sequence ($w_{\textrm{FL}} \to  w_{\textrm{BR}}\to w_{\textrm{FR}} \to w_{\textrm{BL}} \to w_{\textrm{FL}} \to \cdots $) while executing the quasi-static walking. Stability is achieved by controlling the COM to lie above the support polygon of the stance legs and moving slowly enough that the effects of momentum are negligible, hence ``quasi-static''. Based on $h_{\textrm{gait}}(\bm{x}_{\varphi})$, we can make a smooth transition between the gaits as follows:
\begin{equation}
    \Xi = \left\{\begin{array}{ll} \Xi_{\textrm{trot}}(H, \dot{\varphi}_{x}, \dot{\varphi}_{y}, \dot{\vartheta}_{z}) & \textrm{if } h_{\textrm{gait}}(\bm{x}_{\varphi}) \geq 0 \\
    \Xi_{\textrm{static}}(H, \dot{\varphi}_{x}, \dot{\varphi}_{y}, \dot{\vartheta}_{z}) & \textrm{if } h_{\textrm{gait}}(\bm{x}_{\varphi}) < 0 
    \end{array} \right.
\end{equation}
Given that the CBF for ensuring safe locomotion is configured in an ellipsoidal form, it is intuitive to design the CBF for gait transition in a similar elliptical fashion to maintain consistency with the locomotion CBF.

\subsection{Low-level controllers}
We employ a comprehensive approach to compute control commands for the robot, considering its full-body model, reference velocity of the robot's base, and gait type. Drawing inspiration from Raibert's heuristics, we plan the footstep for the swing foot and subsequently solve the inverse kinematics problem to derive the joint configuration corresponding to the planned footstep. During this process, we generate the desired joint position and velocity, denoted as $\bm{q}^{d}$ and $\dot{\bm{q}}^{d}$, respectively. Next, we formulate an optimization problem in the form of a QP to simultaneously determine the desired joint acceleration and the feedforward torque input. This optimization problem takes into account various factors, including the complete full-body model of the robot, contact constraints, constraints imposed by the friction wrench cone, and input limits. The formulation of this optimization problem is outlined below:
\begin{equation}
    \begin{split}
        \min_{\ddot{\bm{q}}, \bm{u}} &\quad (\ddot{\bm{q}}^{d} - \ddot{\bm{q}} )^{\top} \mathbf{W}_{\ddot{q}} (\ddot{\bm{q}}^{d} - \ddot{\bm{q}}) + \bm{u}^{\top} \mathbf{W}_{u} \bm{u} \\
        \textrm{s.t.} &\quad \dot{\bm{x}}_{q} = f_q(\bm{x}_q) + g_q(\bm{x}_q) \bm{u}, \\
        &\quad \psi_{c}(\bm{x}_q, \dot{\bm{x}}_q) = 0, \quad \bm{F}_{c} \in \mathcal{FC},\quad \bm{\tau} \in \mathcal{U},\\
        &\quad \ddot{\bm{q}}^{d} = \mathbf{K}_p(\bm{q}^{d} - \bm{q}) + \mathbf{K}_{d}(\dot{\bm{q}}^{d} - \dot{\bm{q}})
    \end{split}
\end{equation}
where $\mathbf{W}_{\ddot{q}}$ and $\mathbf{W}_{u}$ represent the weighting matrices assigned to the terms associated with joint acceleration error and input, respectively. Furthermore, the desired joint acceleration is determined through the application of a classical Proportional-Derivative (PD) control law, incorporating proportional gains denoted as $\mathbf{K}_p$ and derivative gains denoted as $\mathbf{K}_d$. The computation of desired joint position and velocity is achieved via Euler integration of the optimal solution obtained from the aforementioned optimization problem ($\ddot{\bm{q}}^{\star}$):
\begin{equation}
    \begin{split}
        \dot{\bm{q}}^{n} &= (1-\gamma) \dot{\bm{q}} + \gamma \dot{\bm{q}}_{\textrm{prev}}^{n} + \Delta t \ddot{\bm{q}}^{\star},\\
        \bm{q}^{n} &= \bm{q} + \Delta t((1-\gamma) \dot{\bm{q}} + \gamma \dot{\bm{q}}_{\textrm{prev}}^{n}) +0.5 \Delta t^{2} \ddot{\bm{q}}^{\star}
    \end{split}
\end{equation}
where $0 \leq \gamma \leq 1$ denotes the weighting factor, and $\dot{\bm{q}}_{\textrm{prev}}^{n}$ denotes the previous desired joint velocity to smooth the noisy signal $\dot{\bm{q}}$. The torque command is subsequently computed as follows:
\begin{equation}
    \bm{\tau}_{\textrm{cmd}} = \bm{\tau}^{\star} + \mathbf{K}_{p}^{\textrm{imp}} ( \bm{q}^{n} - \bm{q}) + \mathbf{K}_d^{\textrm{imp}}(\dot{\bm{q}}^{n} - \dot{\bm{q}})
\end{equation}
where $\mathbf{K}_{p}^{\textrm{imp}}$ and $\mathbf{K}_{d}^{\textrm{imp}}$ represent the proportional and derivative gains for the joint impedance controller. In this architecture, the QP is solved at 1 kHz on a control computer, and the joint impedance control is computed on each individual motor driver at 8 kHz.

\begin{figure}[t] 
\centering
\includegraphics[width=\linewidth]{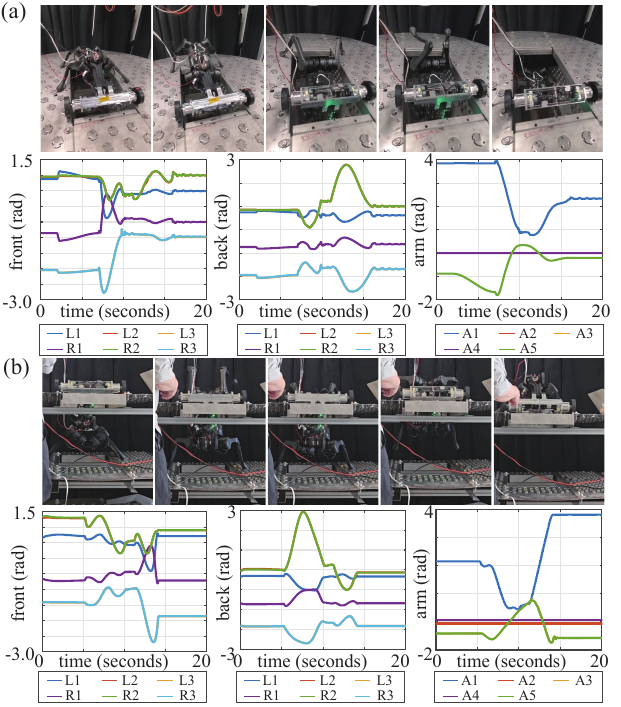}
\vspace{-6mm}
\caption{\textbf{Experiments for transition behaviors}: (a) Downward transition from the upper level to the lower level, (b) Upward transition from the lower level to the upper level. In the labels, "L," "R," and "A" denote the left-side, right-side, and arm joints of the robot, respectively.}
\label{Fig6}
\vspace{-0.7cm}
\end{figure}

\section{Experimental Demonstration}
\label{section4} 

In this section, we present the experimental results showcasing the key functionalities derived from the proposed framework, specifically focusing on safe transitions between levels and safety-critical locomotion with gait transitions. These experiments were carried out using the Unitree A1 quadruped, equipped with $23$ degrees of freedom (DOFs), comprising $18$ DOFs for the robot itself, including virtual joints ($6$ DOFs) describing the robot's floating base, and an additional $5$ DOFs for the roller arm with $4$ actuators and a bidirectional lead screw for its extender. The experiments were conducted on a laptop featuring a 3.4 GHz Intel Core i7 processor. The trajectory generation component was implemented using FROST \cite{hereid2017frost} on MATLAB in an offline manner, while the remaining components were implemented in real-time. We employed OSQP \cite{osqp} as the optimization tool for these real-time operations.

\subsection{Transition behaviors between the levels}
We have employed a multi-layered approach to generate trajectories for both downward and upward transitions within the column. The manway possesses dimensions of $22$ inches ($0.56$ m) in length and $15$ inches ($0.381$ m) in width, with an $18$-inch ($0.458$-m) gap between levels. We consider the constrained sets $\mathcal{Q}$, $\mathcal{V}$, and $\mathcal{U}$ as previously described in \cite{molnar2023mechanical}. The transition duration was uniformly set to 8 seconds for both the downward and upward movements. Through the incorporation of an additional COM constraint and the fine-tuning of parameters, we have successfully demonstrated stable and robust downward (see Fig. \ref{Fig6}(a)) and upward (see Fig. \ref{Fig6}(b)) transitions within the physical column tray. The joint position data displayed in Fig. \ref{Fig6} confirms that these transitions occur without collisions and jerky movements, ensuring safety and stability.

\begin{figure}[t] 
\centering
\includegraphics[width=\linewidth]{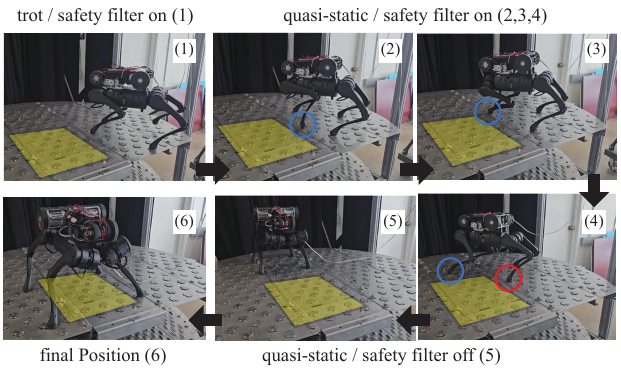}
\vspace{-8mm}
\caption{\textbf{Snapshots of the safety-critical locomotion experiment}: In the confined space, the robot navigates around the manway (indicated by the yellow region) by activating the CBF-based safety filter. Once past the manway, it rotates and proceeds toward it without the safety filter.}
\label{Fig7}
\vspace{-0.7cm}
\end{figure}

\subsection{Safety-critical locomotion with gait transition}
In our second experimental demonstration, we set up a scenario where the robot has to reach a designated goal position on the manway, which serves as the initial position for descending through the manway, as indicated by the yellow rectangle in Fig. \ref{Fig7}. To accomplish this task, the robot must navigate to an intermediate position before proceeding to the goal position on the manway sequentially. This sequential behavior is necessitated by the confined space and limited range of motion, as illustrated in Fig. \ref{Fig7}. The center of the manway is located at ($0.5$, $0$) m relative to the initial base location of the robot ($0$, $0$) m, with dimensions identical to those described in the previous experiment section. The principal axes of the ellipses used for defining $h_{\textrm{path}}$ and $h_{\textrm{gait}}$ are ($a=0.19$, $b=0.31$) and ($a=0.49$, $b=0.88$).

\begin{figure}[t] 
\centering
\includegraphics[width=\linewidth]{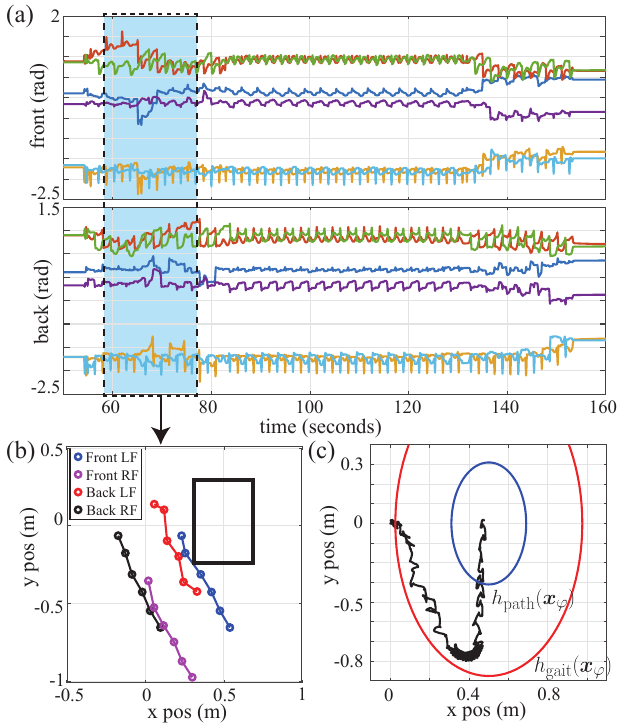}
\vspace{-7mm}
\caption{\textbf{Experimental results of the safety-critical locomotion on the tray}: (a) joint positions of each leg, (b) corresponding footholds (for the blue region from (a)), and (c) the position of the robot's base and the ellipses representing CBFs. Legends in (a) match those in Fig. \ref{Fig6}.  }
\label{Fig8}
\vspace{-0.7cm}
\end{figure}

Fig. \ref{Fig7} provides a detailed account of the robot's behavior throughout the experiment. Initially, the robot commenced diagonal-gait trot locomotion ($\Xi_{\textrm{trot}}$) while employing the safety filter (Fig. \ref{Fig7}-(1)). Upon entering the area enclosed by the red ellipse, representing $h_{\textrm{gait}}(\bm{x}_{\varphi}) <0$ (as illustrated in Fig. \ref{Fig8}(c)), the robot transitions from the trot walking ($\Xi_{\textrm{trot}}$) to the quasi-static walking ($\Xi_{\textrm{static}}$) (Fig. \ref{Fig7}-(2)). As it approaches the intermediate position, the safety of the robot's base is meticulously maintained to avoid the unsafe region denoted by the blue ellipse in Fig. \ref{Fig8}(c). Once the robot reaches the intermediate position, the safety filter is deactivated, allowing the robot to align itself with the manway (Fig. \ref{Fig7}-(5)). Ultimately, the robot successfully reaches the goal position on the manway, as depicted in Fig. \ref{Fig7}-(6).

Throughout the experiment, the robot's footstep planning is adjusted to ensure avoidance of the unsafe region using the proposed re-planning method. This adjustment is evident in Fig. \ref{Fig7}-(2,3,4), where the originally planned locations for the left-side feet are modified to steer clear of the manway. These modifications are prominently illustrated in Fig. \ref{Fig8}(a) and (b). In Fig. \ref{Fig8}(a), the periodic joint positions of the robot's legs are altered to ensure safe footholds, indicated by the blue section in Fig. \ref{Fig8}(a). Despite the initial planned footsteps being within the manway, the left-sided foot positions remain within the safe region, as evidenced in Fig. \ref{Fig8}(b). This experiment substantiates the capability of our proposed framework to ensure safety in both reduced-order and full-order systems.

\section{Conclusion}

This paper introduces an innovative layered architecture tailored for a quadruped robot equipped with a roller arm, intended for operation within layered environments. We enhance the trajectory generation method to facilitate smooth transitions between levels and substantiate its efficacy through hardware experiments. We ensure safety in both reduced-order and full-order systems by leveraging a CBF-based safety filter and our footstep re-planning method. Moreover, the robot autonomously selects gaits based on the assessed level of danger. The framework is verified by demonstrating the safety-critical locomotion in the confined space, which is an actual industry-grade column tray.

In the near future, we intend to augment our framework by incorporating perception and localization components to mitigate uncertainties stemming from the model-based approach. Furthermore, we envisage deploying this robotic system with full autonomy, governed by a state-machine, to undertake comprehensive inspection missions within multi-tiered environments.

\bibliographystyle{IEEEtran}
\bibliography{l_css}

\end{document}